\documentclass[letter, 10pt, conference]{ieeeconf}
\IEEEoverridecommandlockouts
\title{\LARGE \bf State Representations as Incentives for Reinforcement Learning Agents: A Sim2Real Analysis on Robotic Grasping}

\author{Panagiotis Petropoulakis$^{1\ast}$, Ludwig Gr{\"a}f$^{1\ast}$, Mohammadhossein Malmir$^{1\dag}$, Josip Josifovski$^{1\dag}$, and Alois Knoll$^{1}$
\thanks{This work has been financially supported by the A-IQ READY project, which has received funding within the Chips Joint Undertaking (Chips JU) - the Public-Private Partnership for research, development, and innovation under Horizon Europe – and National Authorities under grant agreement No. 101096658.}
\thanks{$^{1}$ The authors are with the Department of Computer Engineering, School of Computation, Information and Technology, Technical University of Munich, Germany.}%
\thanks{$^{\ast}$ Panagiotis Petropoulakis and Ludwig Gr{\"a}f are co-first authors.}%
\thanks{$^{\dag}$ Mohammadhossein Malmir and Josip Josifovski are co-second authors.}}%

\usepackage{cite}  
\usepackage{amsmath,amssymb,amsfonts}
\usepackage{algorithmic}
\usepackage{graphicx}
\usepackage{textcomp}
\usepackage{xcolor}
\usepackage{xfrac}
\usepackage[acronym]{glossaries}
\usepackage{subfig}
\usepackage{gensymb}

\usepackage{url}
\usepackage[protrusion=true,expansion=true]{microtype}
\usepackage{hyperref}

\hypersetup{
  pdftitle={State Representations as Incentives for Reinforcement Learning Agents: A Sim2Real Analysis on Robotic Grasping},
  pdfauthor={Panagiotis Petropoulakis, Ludwig Gr{\"a}f, Mohammadhossein Malmir, Josip Josifovski, and Alois Knoll},
  pdfsubject={Robotics (cs.RO); Artificial Intelligence (cs.AI)},
  pdfkeywords={Deep Reinforcement Learning, State Representation Learning, Autoencoders, Incentives, End-to-End Learning, Real-World, Sim-to-Real, Robotic Grasping, Pre-Trained Models},%
  colorlinks=true,
  linktoc=all,
  linkbordercolor=white,
  linkcolor=black,
  citecolor=black,
  urlcolor=black,
}
\newcommand{\specialcell}[2][c]{%
  \begin{tabular}[#1]{@{}c@{}}#2\end{tabular}}

\newacronym{DL}{DL}{Deep Learning}
\newacronym{CV}{CV}{Computer Vision}
\newacronym{NN}{NN}{Neural Networks}
\newacronym{DNN}{DNN}{Deep Neural Networks}
\newacronym{RL}{RL}{Reinforcement Learning}
\newacronym{CNN}{CNN}{Convolutional Neural Network}
\newacronym{MDP}{MDP}{Markov Decision Process}
\newacronym{MLP}{MLP}{Multi-Layer Perceptron}
\newacronym{ROS}{ROS}{Robot Operating System}
\newacronym{AE}{AE}{Autoencoder}
\newacronym{IGAE}{IGAE}{Incentivized Grasping AutoEncoder}
\newacronym{MSE}{MSE}{Mean Squared Error}
\newacronym{CC-VAE}{CC-VAE}{Context-Conditional Variational Autoencoder}
\newacronym{PPO}{PPO}{Proximal Policy Optimization}
\newacronym{VtS}{VtS}{Vision-to-State}
\newacronym{DoF}{DoF}{Degrees of Freedom}
\newacronym{MLE}{MLE}{Maximum Likelihood Estimation}
\newacronym{BCE}{BCE}{Binary Cross-Entropy}
\begin{document}
\maketitle
\thispagestyle{empty}
\pagestyle{empty}

\begin{abstract}
Choosing an appropriate representation of the environment for the underlying decision-making process of the reinforcement learning agent is not always straightforward. The state representation should be inclusive enough to allow the agent to informatively decide on its actions and disentangled enough to simplify policy training and the corresponding sim2real transfer. Given this outlook, this work examines the effect of various representations in incentivizing the agent to solve a specific robotic task: antipodal and planar object grasping. A continuum of state representations is defined, starting from hand-crafted numerical states to encoded image-based representations, with decreasing levels of induced task-specific knowledge. The effects of each representation on the ability of the agent to solve the task in simulation and the transferability of the learned policy to the real robot are examined and compared against a model-based approach with complete system knowledge. The results show that reinforcement learning agents using numerical states can perform on par with non-learning baselines. Furthermore, we find that agents using image-based representations from pre-trained environment embedding vectors perform better than end-to-end trained agents, and hypothesize that separation of representation learning from reinforcement learning can benefit sim2real transfer. Finally, we conclude that incentivizing the state representation with task-specific knowledge facilitates faster convergence for agent training and increases success rates in sim2real robot control.\footnote[2]{Supplementary materials can be found on the project webpage: https://github.com/PetropoulakisPanagiotis/igae}
\end{abstract}

\section{Introduction}

The ultimate goal of intelligent agents is to maximize their utility by learning to command resolute actions while relying on observations that can serve as incentives for events \cite{everitt2022understanding}. In \gls{RL}, the agent's utility is closely associated with its return, which is the expectation of the cumulative sum of the rewards received at every step of the perception-action cycle along the temporal planning horizon \cite{sutton2018reinforcement}. To obtain the best possible return, the \gls{RL} agent requires an appropriate representation of its environment state.

If we consider state representations in \gls{RL} for robot control, the representation space is usually partitioned into two sub-spaces: the Vision State Space where the agent receives raw image data, and the Numerical State Space where the agent relies on hand-crafted states (e.g., derived from the robot encoders or pre-processed raw sensor data) suitable for the task. While the raw image data is a more general representation, it is often not directly useful as-is, since it contains unprocessed and dense information that could hinder the learning process. For example, a significant performance gap was evident in the DeepMind Control Suite \cite{deepmind-suite} between numerical and vision-based states, where the agents trained directly on images required, on average, more than 60 million steps to reach convergence, while harder tasks could not be solved at the same level compared to the agents that relied on hand-crafted numerical states.

The influence of learning pre-trained representations on the development of decision-making agents cannot be overstated \cite{bengiorepr}. State representations play a pivotal role in enhancing the efficiency of \gls{RL} agents throughout the policy learning process, introducing inductive biases into the features, and accelerating exploration \cite{NEURIPS2022_0fd489e5}. It can be argued that it is through the constraints imposed by the representations that agents formulate their incentives to learn the assigned task. In each task, segments of the input information that hold positive values in helping the agent make better decisions (e.g., for optimizing its policy) are referred to as observation incentives or the value of information \cite{Everitt_Carey_Langlois_Ortega_Legg_2021}.

Consequently, the creation of concise representations as observation incentives necessitates the identification and selection of the most essential and informative elements available from the environment. To do this, intervention incentives are used to influence observation incentives by changing the conditional probability distribution of the node \cite{everitt2022understanding} and, hopefully, enriching the representation and accelerating exploration \cite{NEURIPS2022_0fd489e5}. Intervention incentives can be, for example, additional objectives for the task, designing specific neural network architectures, or adding randomization noise to the observations. 

Many works have addressed the problem of shaping an appropriate state representation for \gls{RL} agents in robotics. However, there is rarely a comparison of how different state representations would influence the control performance of the agent or, even more importantly, how the representation influences the transferability of the model to the real system in the context of simulation-based robot learning. To provide some insight into these questions, in this work we consider a continuum of state representations for a robotic grasping task, ranging from a hand-crafted numerical state to an abstract representation that should be learned from raw image data. An objective comparison has been made to assess which of the approaches are more suitable based on the specific requirements of the task, available system and environment information, and the level of robust transferability required for real-world applications. Our analysis shows that when high performance and smooth sim2real transfer are necessary, general representations that do not explicitly incentivize the available domain knowledge are underperforming. To address this problem, we propose the \gls{IGAE} architecture, an image-based approach that introduces domain-specific knowledge and enhances the agent's performance while holding a general representation.

\section{Background and Related Work}

When one defines the state representation for the \gls{RL} agent, this definition might be more tailored to the specific task and reward-driven\cite{DBLP:conf/icml/DannMMSS22}, or it can be reward-free and involve unsupervised, self-supervised, or pre-trained models to create more generic features. For reward-free methods, there are numerous paradigms to extract meaningful representations. In \cite{flambe}, the authors examined how \gls{MLE} can approximate the dynamics of data distributions that originate from exploratory policies, ultimately leading to the learning of a low-dimensional feature representation. Other approaches \cite{uehara2022representation, Zhang2022EfficientRL} involve the utilization of adversarial losses, where a minimax objective is formulated, after which standard gradient-based methods are employed to optimize the discriminators' functions responsible for generating the state representations. 

To address the challenge of data efficiency and generalization to new environments in both vision-based and hand-crafted numerical representations, recent efforts have introduced data augmentation techniques borrowed from the computer vision domain \cite{deepmind-suite}. The Reinforcement Learning with Augmented Data (RAD) module \cite{10.5555/3495724.3497393} showed that even basic randomization added to state representations alone outperforms more complex methods, such as pixel SAC \cite{pmlr-v80-haarnoja18b}.

While randomization can improve the final performance, in some cases it can also increase the complexity of the task learning and hinder the optimization process. In \cite{josifovski2022analysis}, the effects of randomization in numerical-based agents for the sim2real transfer of a robotic reaching task were analyzed. The results showed that randomization helps in sim2real transfer, but inappropriate randomization ranges can prevent the agent from finding a good policy.

Recently, several works have focused on decoupling the perception module from the policy and reducing the computational overhead of \gls{RL} agent training, intending to address more complex tasks. Parisi et al. \cite{Parisi2022TheUE} demonstrated the effectiveness of using general pre-trained feature embeddings from ImageNet \cite{deng2009imagenet} to train a shallow \gls{MLP} policy. This approach successfully solved several tasks, where an end-to-end visuo-motor policy failed. In \cite{multi-view-grasp}, the authors combined multi-view image observations and visuo-motor feedback through a Grasp Q-Network. The state representation of the \gls{RL} agent involved a latent feature vector derived from convolutional and fully connected layers. This representation enabled a 7-DOF Baxter robot to achieve a grasping accuracy of over 90\% for specific object categories.

In contrast to the impressive results, none of the above works have analyzed the confining effect of the learned representations on transferring the simulation-trained policies to real-world setups. A successful sim2real transfer partly emerges from state representations that feature robustness to the distributional shift raised by the domain gap. Due to their unsupervised nature, autoencoders have emerged as a key architecture for implementing the \gls{RL} perception module for several robotic tasks in sim2real domains.

Breyer et al. \cite{Breyer2018FlexibleRG} presented an autoencoder-based approach to map a masked input image of an object of interest to a low-dimensional vector and defined an efficient and dense reward function with a prioritized \gls{RL} sampling scheme to curate difficult cases during training. This approach achieved high success rates in tasks involving picking and lifting objects. In the work by Nair et al. \cite{nair19ccrig}, a \gls{CC-VAE} network was trained while taking both an input condition, the goal image observation, and the image of the current state observation into account. The reward signal for the \gls{RL} agent was then directly defined within this latent space. This formulation allowed users to specify the goal condition (image) during testing, enabling the robotic system to push an object to the desired position. Similarly, Zhan et al. \cite{zhan2022learning} utilized learned latent vectors extracted from an autoencoder trained in a contrastive manner. They combined several data augmentation techniques and demonstration paradigms and then optimized a policy in just 30 minutes to control a manipulator capable of performing diverse tasks, e.g., pulling large objects and opening drawers.

Unlike the earlier works above, we are showing an empirical evaluation of how various state representations with distinct abstraction levels incentivize the agent learning process and mainly focus on analyzing which representations can attain higher robustness for zero-shot sim2real transfer.

\section{Methodology}

In this section, we start by formally defining a task suitable for comparing different state representations. We then explain in detail the image processing architecture for the vision-based representations, the randomization as an intervention incentive for sim2real transfer, and finally, the evaluation process. 

\subsection{Task Formulation}

The task is to learn a control policy for a robotic manipulator to perform an antipodal grasp of an object placed at a random position in its workspace. The manipulator has seven \gls{DoF} and is controlled in the task space. Conventionally to dynamic planar grasping, the \gls{RL} agent learns to command the actions as specified by 
\begin{equation}
a_t = [v_{x_{t}}, v_{y_{t}}] \in \mathbb{R}^{2} .
\end{equation}

After the agent completes each \gls{RL} episode (i.e., after passing a constant amount of $N$ timesteps per episode), predefined actions are applied to attempt grasping. Namely, if there are no collisions, the following terminal actions are manually initiated at the conclusion of the episode: 1) move the end effector down to a predetermined height, 2) close the gripper, and 3) move the end effector up to a predetermined height. Therefore, successful grasping can only occur if the end effector is in the ideal position before the start of the terminal actions. If a collision occurs at any point, a penalty is added once in the total return, and the episode is considered unsuccessful. This penalty is applied only when: 1) there is a collision of the fingers with the object, 2) there is a collision of the robot with itself, or 3) the safety limit of the joints is violated. For the remaining steps of the episode after the collision, the \gls{RL} agent receives zero rewards, with zero velocities commanded to the manipulator.

More formally, the reward function that guides the agent in learning the task is structured as
\begin{equation}
 r_t =
  \begin{cases}
   w_x * \Delta{d}_{x_{t}} + w_y * \Delta{d}_{y_{t}}, & \text{no collision} \\
   r_c, & \text{collision}
  \end{cases} ,
\end{equation}
where $t$ ranging from 0 to $N$ represents the episode timestep, with $N$ denoting the episode duration. $\Delta{d}_{x_t}$ and $\Delta{d}_{y_t}$ are the normalized differences between the distances from the end effector to the object in the $x$ and $y$-axes at the current timestep compared to the previous one. In particular, $\Delta{d}_{x_{t}}$ is equivalent to $\left(d_{x_{t-1}} - d_{x_{t}}\right)$. Here, $d_{x_{t}} $ is the distance of the end effector to the object at timestep $t$, and $d_{x_{t-1}} $ is the distance at $t-1$, the previous timestep\footnote[3]{In the analysis, we found that the best performance was achieved by separating the errors of the two axes into distinct terms. The conventional practice of using the negative Euclidean distance from the end effector to the object as a reward for grasping led to significantly slower convergence than our modification.}.

\subsection{Image Processing Backbone}\label{backbone}

\begin{figure}[h]
    \centering
    \includegraphics[width=0.48\textwidth]{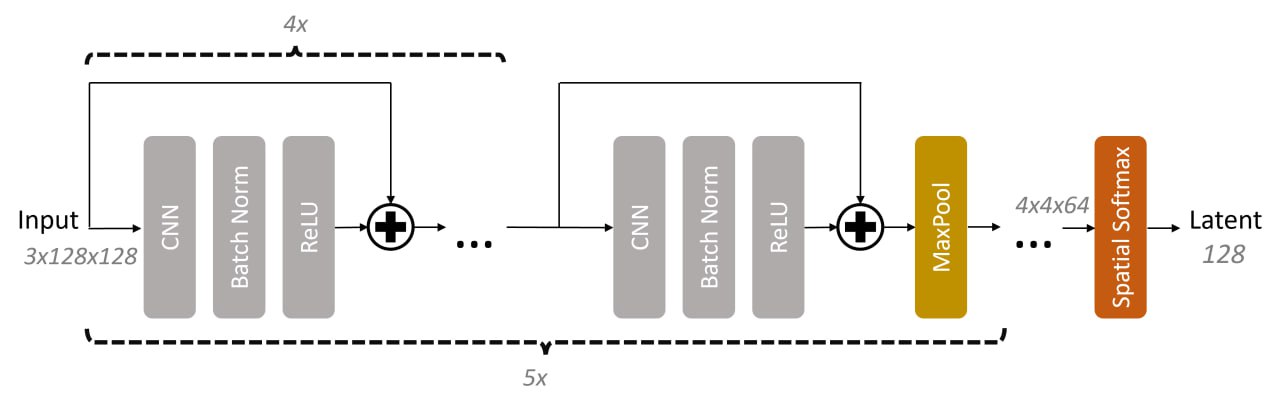}
    \caption{The same backbone architecture was consistently used in all the vision-based \gls{RL} agents. This decision was made to assess the impact of various training objectives in shaping the original image space into meaningful representations of the environment. The architecture itself is an adaptation of the ResNet \gls{AE} \cite{wickramasinghe2021resnet} with a spatial SoftMax output layer \cite{levine2016end, Russ2019}.}
    \label{fig: Encoder Architecture}
\end{figure}

To examine vision-based approaches and mitigate the high training time disparities compared to the numerical agents, while also ensuring fair comparisons across methods, image observations are pre-processed by the same encoder, as depicted in Figure \ref{fig: Encoder Architecture}. The output then represents the input state for the \gls{RL} agent. Nevertheless, to create distinct state representations, we introduce variations in the training objectives. This approach allows us to gain deeper insights into how intervention incentives can transform the original space into a more meaningful and sample-efficient space for the policy to learn the task.

\subsection{State Representations Continuum}

\begin{figure*}[t!]
    \centering
    \includegraphics[width=0.93\textwidth, height=4.5cm]{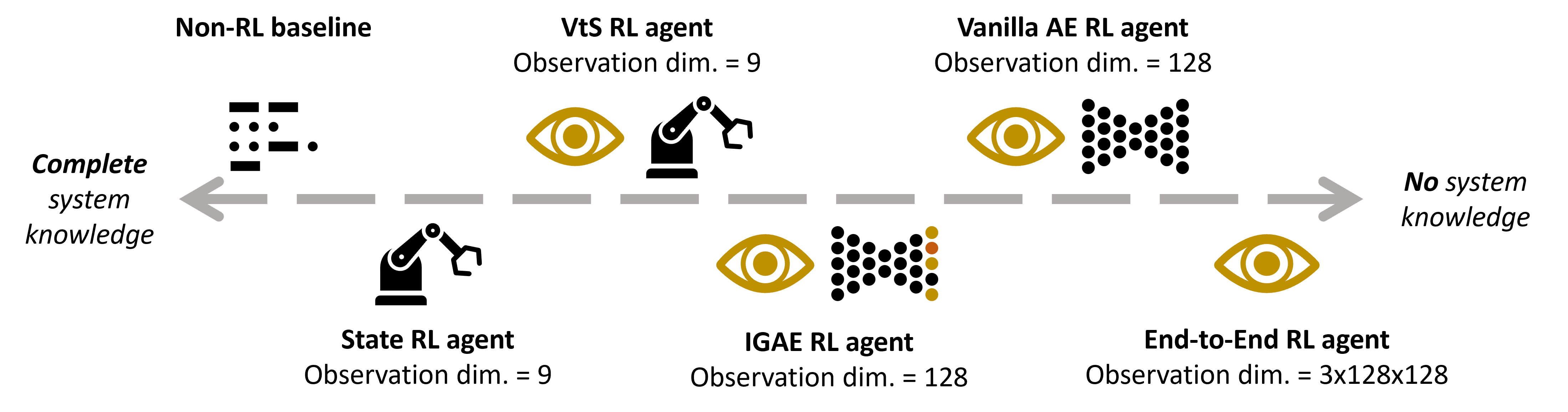}
    \caption{State Representations Continuum: A model-based baseline with full system knowledge and multiple \gls{RL} approaches having different state representation spaces with decreasing levels of system knowledge (increasing abstraction) are examined. Models on the left have the state information explicitly available in the form of numerical values for the robot joints or the object, whereas vision-centric techniques on the right generate implicit latent representation to guide the \gls{RL} policy.}
    \label{fig: Comparison}
\end{figure*}

Six state representation spaces are formulated in this and the next section to train \gls{RL} agents through the reward function and compared with a non-\gls{RL} baseline. The methods vary from predefined hard-coded spaces to a more general end-to-end paradigm, as shown in Figure \ref{fig: Comparison}.

\textbf{Non-\gls{RL} Baseline}: The baseline method employed is Ruckig, a real-time, open-source, and time-optimal trajectory generation algorithm described in \cite{berscheid2021jerk}. The baseline has full knowledge of environment dynamics, and therefore, it serves as an upper bound for the performance of the \gls{RL} agents.

\textbf{State \gls{RL} Agent}: The first learning agent and the simplest state representation is implemented through a hard-coded numerical approach. This particular agent, denoted as the State \gls{RL} agent\label{state-numerical}, relies on the manipulator's encoder measurements, and its state space is defined as

\begin{equation}
s_t = [{\Delta}x_t, {\Delta}y_t, q_{1_t}, ..., q_{7_t}] \in \mathbb{R}^{9} ,
\end{equation}
where ${\Delta}x_t$ and ${\Delta}y_t$ are the positional errors between the end effector and the target position of the object in x and y-axes, and $q_{i_t}$ is the i-th joint position of the manipulator.

\textbf{Vision-to-State \gls{RL} Agent}: The \gls{VtS} \gls{RL} agent\label{vision-state} represents the transition between numerical and image-based representations. Rather than relying on the manipulator's encoders, the joint angles and the positional errors, as described above, are predicted here from image observations. The convolutional backbone (see section \ref{backbone}) processes RGB images and a shallow \gls{MLP} consisting of 4 layers is pre-trained to predict the numerical state of the agent. Due to the occlusion and finite training data, a perfect numerical state is not observable from the image space and the output of MLP contains regression error. Hence, to keep the comparison fair, an identical state RL agent is entirely trained over the estimated states.

\begin{figure}[h]
    \centering
    \includegraphics[width=0.42\textwidth]{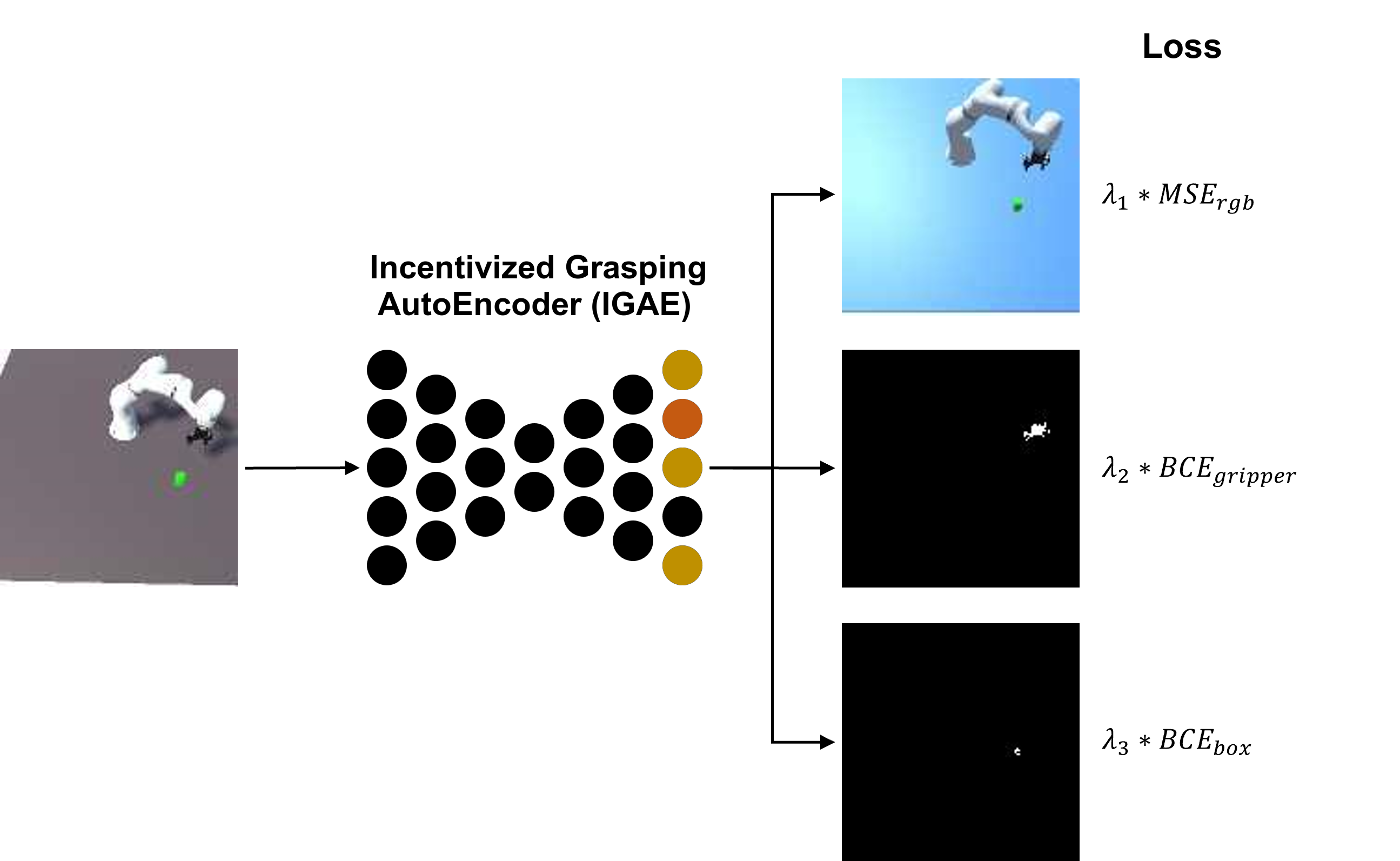}
    \caption{The Incentivized Grasping AutoEncoder (IGAE) takes an augmented image as input and performs reconstruction tasks. It reconstructs first the original (denoised) RGB image, then the gripper, and finally the object binary masks. The total loss is a weighted contribution of each reconstruction loss through $\lambda_i$ terms. In our settings, $\lambda_1$ is set to 1, $\lambda_2$ to 10, and $\lambda_3$ to 20.}
    \label{fig: MAE}
\end{figure}

\textbf{Incentivized Grasping AutoEncoder \gls{RL} Agent}: The next level of state representation is an autoencoder-based latent representation derived from image observations. In AutoEncoder RL agents, the trained latent vector is frozen and then processed by the actor-critic networks for the agent training phase. To achieve a balanced blend of general representation and task-specific precision, we introduce the Incentivized Grasping AutoEncoder (IGAE) architecture (see Figure \ref{fig: MAE}). The IGAE is a segmentation-inspired \gls{RL} agent tailored for the grasping task by introducing additional objectives, i.e. intervention observation incentives. These incentives are designed with the intention of isolating and shaping specific regions of the vision space, guiding the \gls{RL} agent to learn the task in a more sample-efficient manner.

\textbf{Vanilla AutoEncoder \gls{RL} Agent}: In this level of state representation, we are now progressively eliminating human-injected knowledge, and shifting to a more abstract and generic representation. For the vanilla AE agent, we discard the extra objectives introduced in the \gls{IGAE} agent and retain only the RGB reconstruction head to project the vision space into a more generic latent feature representation. 

\textbf{End-to-End \gls{RL} Agent}: In the most extreme case of defining a representation space, we grant complete freedom to the \gls{RL} policy. We couple representation learning with the policy learning step and update the vision module's weights as the agent learns the grasping task. The primary goal of the end-to-end \gls{RL} agent is to learn the task relying solely on the reward signal by removing all intervention observation incentives introduced earlier. This increases the number of parameters to be learned, potentially resulting in slower training, reduced sample efficiency, and higher complexity in the \gls{RL} algorithm.

\subsection{Domain Randomization for Sim-to-Real Transfer}
\label{sec: Domain Randomization for Sim-to-real Transfer}

Training \gls{RL} agents in idealized simulation environments is insufficient to transfer their policy into real-world systems due to the sim2real gap \cite{sim-to-real-survey}, \cite{malmir2020robust}. Simulations can not accurately capture the expected noise of the robot's measurements nor the exact appearance, for instance, lighting conditions of real robotic setups.

In the scope of our analysis, we focus on state representations that incentivize simulation-trained \gls{RL} agents to solve the task both in simulation and in the real system. We adopt the core ideas from the RAD module \cite{10.5555/3495724.3497393} and Tobin et al. \cite{state-predictor}, which demonstrated successful agent transfer via domain randomization in both numerical and vision-based domains. 

For all the image-based models, domain randomization is applied in two distinct ways: 1) by directly adding noise to the images generated by the simulator, and 2) by first applying physical adaptations to the simulation environment and then introducing additional noise through step 1.  

In the first step, the following augmentations are randomly applied: Random Cropping, Gaussian Blurring, Random Changes in Brightness, Random Four-Point Perspective Transform, and Random Channel-Wise Noise. Finally, for the second step, our simulator\footnote[4]{The simulator used for the experiments is available at: https://github.com/tum-i6/VTPRL} offers a wide range of physical adaptations to enhance realism and variability: Random Floor Appearance, Random Texture, Shadow Randomization, and Random Camera Pose Perturbations.

\textbf{State Randomized \gls{RL} Agent}: In the case of numeric observations, a more conservative approach is followed since the manipulator's encoder measurements are less noisy than the images. This agent is similar to the State \gls{RL} Agent, with the difference that a uniform noise of $5\%$ is applied at every timestep $t$ in the state of the agent during the policy training. The magnitude of the random noise resulted from optimization over preliminary sim2real experiments in order to create a non-jittering motion.

\begin{figure}[h]
    \centering
    \includegraphics[width=0.39\textwidth]{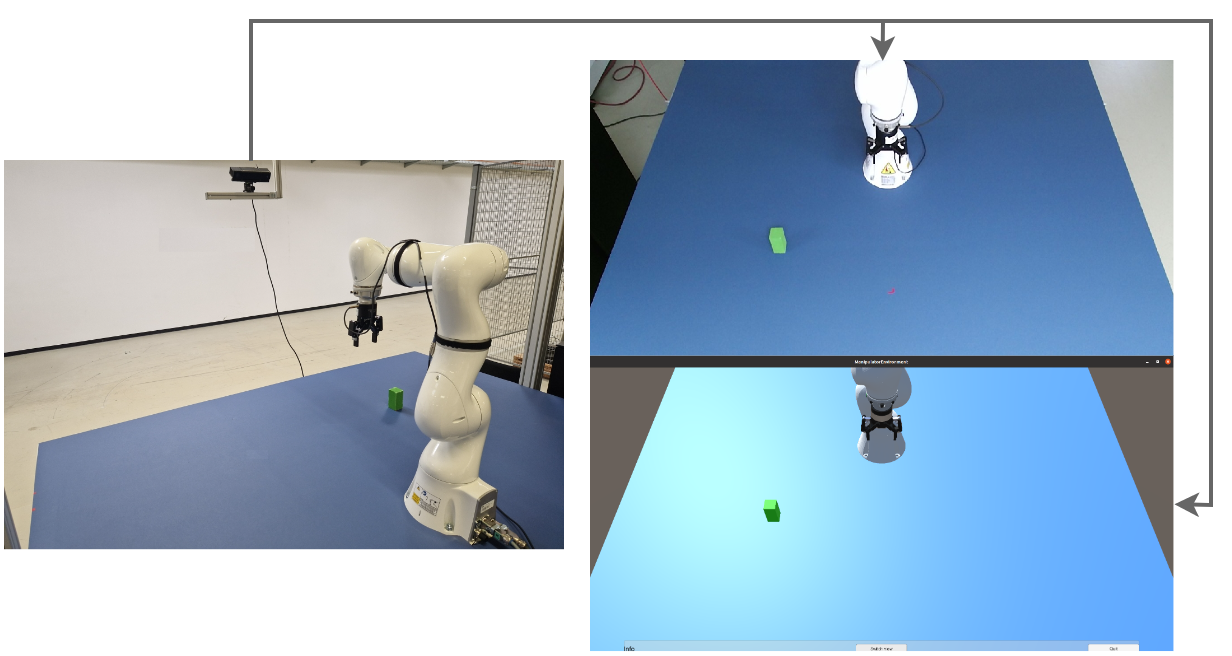}
    \caption{Overview of the real-world setup (left). Image observation from the Kinect V2 camera (top), and the aligned image observation in the simulated environment (bottom).}
    \label{fig: simulation}
\end{figure}

\section{Experiment Setup}

\subsection{Environment Description}

For the experiments, we use the VTPRL simulation environment introduced by \cite{Josifovski2020ContinualLO}, developed with the Unity game engine \cite{Unity} and coupled with the Dynamic Animation and Robotics Toolkit (DART) \cite{Lee2018} for the inverse kinematics calculations. It has a comparable API to OpenAI Gym \cite{Brockman2016OpenAI} and allows parallel simulation of several robots to speed up the training process. We use the KUKA LBR iiwa 14 robotic manipulator \cite{Kuka} and the 2-Finger Adaptive Robotic Gripper (2F-85 model) by Robotiq \cite{gripper}, with a maximal opening of 85 mm. The simulated manipulator and gripper parameters and meshes are obtained from the URDF data provided by the ROS-Industrial \cite{ROSindustrial} and Robotiq \cite{robotiqgit} packages. The real robot is controlled with the IIWA stack \cite{Hennersperger2017MRIBased} via \gls{ROS} \cite{ros-operating-system}. In the real setup, the image observations are taken with the Microsoft Kinect V2 camera sensor \cite{fankhauser2015kinect} and processed with the iai\_kinect2 \cite{iai_kinect2} package. The object we grasp is a foam box with dimensions of 50 x 100 x 50 mm (width x height x depth).

\subsection{Implementation Details}

\textbf{AutoEncoders}: The agent learning process here is divided into three separate steps. 1) Training images and segmentation masks are initially generated from the simulation based on the unique color of the gripper and the target object. 2) The autoencoders are then trained to compress augmented (noisy) images into 128-dimensional latent vectors. This is achieved by reconstructing the original (denoised) RGB images, guided by an \gls{MSE} loss objective \cite{joshi2021deepurl}. As a result, a simulated and a corresponding real-world image are encoded to similar latent representations. Additionally, to further shape the latent space, the \gls{IGAE} predicts two binary segmentation masks with \gls{BCE} loss objectives, one for the gripper and one for the target object. 3) Once the autoencoders have been optimized, the frozen encoders process the image observations, and the agents learn a policy through interactions with the environment while relying on the latent vector as a representation of the environment states.

\textbf{\gls{RL} Policy}: For all representation spaces we used the model-free algorithm \gls{PPO} \cite{PPO} to train the agents. This algorithm is particularly well-suited for continuous action spaces, and the novel proposed clipping coefficient can effectively address the high variance issues in the policy update step. We used the parallel PPO implementation from StableBaselines3 \cite{stable-baselines3}. 

Regarding the architecture of the policy network, the default implementation for the PPO algorithm of the StableBaselines3 (2 fully connected layers of 64 units and Tanh activation functions for each of the actor and the critic's heads) showed the best performance. However, in the vision-based methods, replacing the Tanh with ReLU activation functions yielded more stable results\footnote[5]{Most likely, the sparsity property contributed to the improved behavior, given the larger input dimension compared to the State \gls{RL} agent. We also observed that other types of activation functions performed much worse, and smaller policy networks always outperformed larger ones in our setup.}. 

The hyperparameters for the policies and autoencoder architectures were determined using the Optuna Search framework \cite{optuna_2019}. The policy optimization process was carried out on the State \gls{RL} agent, and the best-found hyperparameters were also applied to the other agents. However, the learning rate $(\alpha)$ and clip range $(\epsilon)$ were optimized from the beginning in all approaches since these parameters are highly sensitive to the dimension of the representation, ensuring fair comparisons across the methods. The hyperparameter settings for PPO can be found in Tables \ref{table:hyperparameters_details_individual} and \ref{table:hyperparameters_details_shared}.

\begin{table}[t]
    \begin{center}
    \caption{Representation-specific PPO hyperparameters.}
    \label{table:hyperparameters_details_individual}
    \begin{tabular}{lc|c|c|c}
& St. & St. (rnd.) \& VtS & \gls{AE} \& \gls{IGAE} & EtE \\
    \noalign{\smallskip}\hline\noalign{\smallskip}
     LR ($\alpha$)                       & 0.001 & 0.0005 & 0.0001 & 0.00005\\
     Clip range ($\epsilon$)                        &  0.3 & 0.25 & 0.25 & 0.25\\ 
    \noalign{\smallskip}\hline
    \end{tabular}
    \end{center}
\end{table}

\begin{table}[t]
    \begin{center}
    \caption{Representation-invariant PPO hyperparameters.}
    \label{table:hyperparameters_details_shared}
    \begin{tabular}{lc}
    \hline\noalign{\smallskip}
     Entropy coefficient ($c_{2}$)                  & 0.02 \\
     Discount factor ($\gamma$)                     &  0.96 \\ 
     Bias-variance trade-off ($\lambda$)            &  0.92 \\
     Number of steps ($n\_steps$)                   &  512 \\
     Number of optimization epochs ($n\_epochs$)    &  10 \\
     Value function coefficient ($c_{1}$)           &  0.5 \\
     Gradient Clipping ($max\_grad\_norm$)          &  0.5 \\
     Mini-batches ($batch\_size$)                   &  32 \\ 
    \hline\noalign{\smallskip}
    \end{tabular}
    \end{center}
\end{table}

The agents' actions and states are scaled to fall within the range of $[-1, 1]$ for stable training of the policy. The maximum task-space velocities are also set to $0.035 m/sec$, and the configured control cycle is $20Hz$. For the non-learning baseline, we intentionally remove the acceleration and jerk constraints to ensure a fair alignment with the \gls{RL} definitions. In total, the complete 6-\gls{DoF} Twist vector of the robot end effector is controlled, and only the two planar velocities are commanded as $a_t$ by the trained agents. Hence, to force the robotic manipulator to perform planar movements and maintain a consistent pose across its uncontrolled \gls{DoF} (4 in total), a Proportional (P)-controller is utilized. The gains are $0.5$, and $1.5$ for the positional and rotational errors, respectively. The same P-controller is employed across all different approaches.

The episode duration is uniform and set to $400$ timesteps, and the \gls{RL} agents were trained for a total of $1.5$ million timesteps. The reward weights $w_x$ and $w_y$ are both set to a value of $1.0$ and the collision term $r_c$ to a value of $-100$. The joint limits are set to $-10^{\circ}$ from the physical limits. We finally trained on 16 environments in parallel, and for evaluation, $50$ random object positions were selected to assess the $30$ saved model checkpoints.

\section{Experiment Results}

\begin{figure}[t]
    \includegraphics[width=0.48\textwidth]{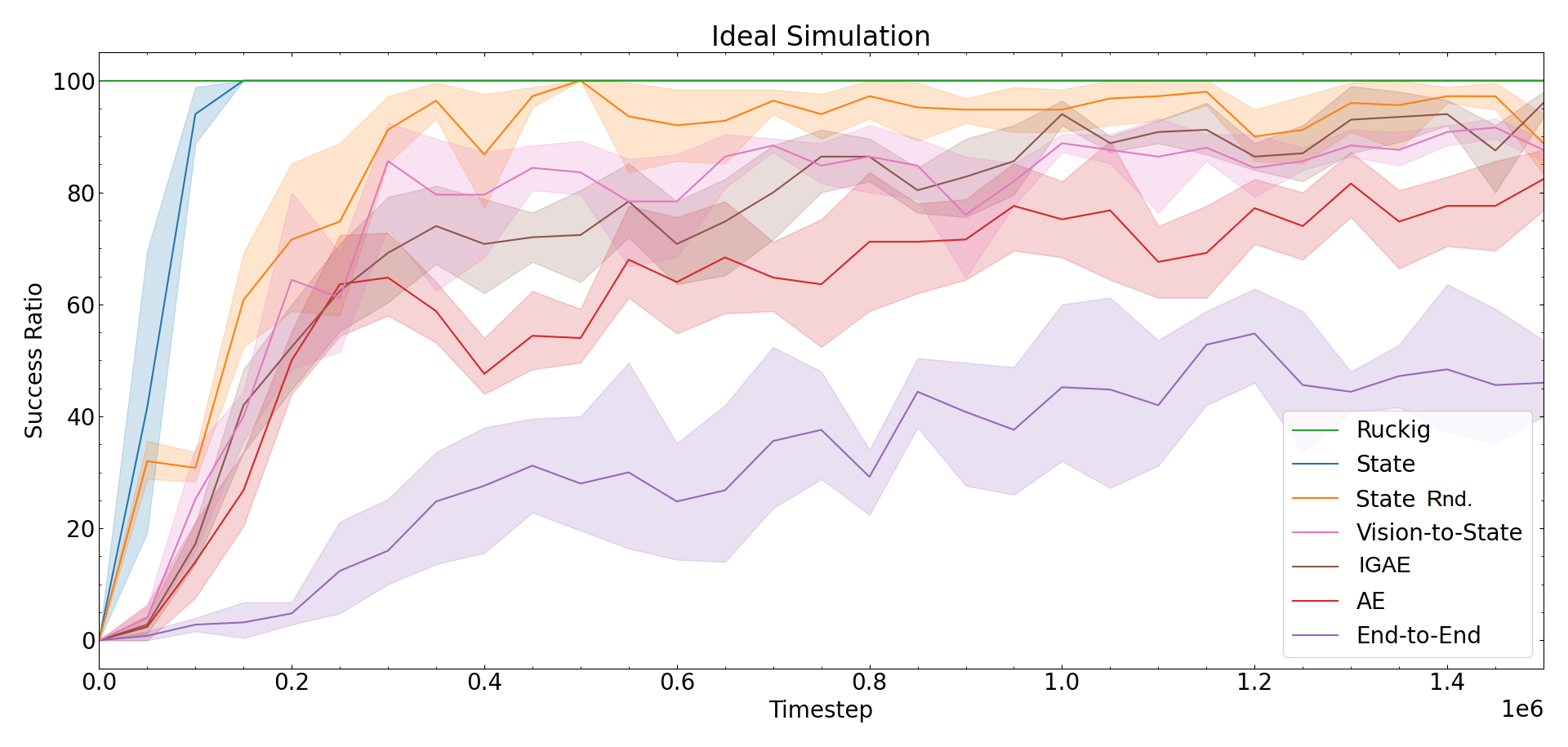}
    \caption{The success rate development in training the proposed agents, each utilizing distinct representation space.}
    \label{fig: succ-ratio-ideal}
\end{figure}

The success rate development during the training of the \gls{RL} agents, along with the non-learning baseline on the $50$ evaluation object positions, is shown in Figure \ref{fig: succ-ratio-ideal}. These results consist of an average of 5 runs with distinct initialization seeds. The grasping task can be completely solved with a 100\% success rate using the following methods: 1) Ruckig, 2) Ideal State, 3) Randomized State, and 4) IGAE \gls{RL} agents.

Most approaches exhibit small standard deviations, apart from the vanilla autoencoder and end-to-end \gls{RL} agents. In the Ideal State \gls{RL} agent, after 180.000 timesteps, the standard deviation across all five runs becomes zero. In the absence of errors and noise, numerical \gls{RL} agents can solve the task as well as non-learning control methods that have full system knowledge. However, introducing only 5\% of uniform random noise in the observation, i.e. the Randomized State \gls{RL} agent, results in a 2x slower convergence rate.

The task-specific IGAE \gls{RL} agent is the best-performing vision-based method, with the best-trained model showing a 100\% success rate (Table \ref{table:experiment2Results}). We hypothesize that this is due to the additional object and end-effector reconstruction heads, which induce a rich latent feature vector that can effectively incentivize the policy to understand the important parts of the environment. The \gls{VtS} and vanilla \gls{AE} agents then follow with success rates of 94\% and 92\%, respectively. However, when the best models are evaluated in a simulation environment with heavy random physical adaptations applied (e.g., camera position), the performance drops to 78\% for the IGAE and 70\% for both \gls{VtS} and vanilla AE (Table \ref{table:experiment2Results} third column).

During the sim2real transfer, we evaluate the checkpoints for the vision-based models that perform best in randomized simulations to gain insights into the effectiveness of the proposed augmentations in bridging the sim2real gap.

On the real system, the Ruckig baseline along with both the Ideal and Randomized State \gls{RL} trained models, consistently achieved successful picking of all 50 evaluation objects. The top-performing vision-based agent on the real setup is the \gls{IGAE} \gls{RL} agent, with a success rate of 84\%. In contrast, all other agents relying on image observations experienced a significant drop in their ability to grasp the targets. Specifically, the vanilla autoencoder achieved a success rate of 60\%, followed by the \gls{VtS} with 52\%. The end-to-end paradigm exhibited the lowest success rate at 24\%.

In qualitative terms, as is evident in the supplementary video\footnote[6]{The supplementary video can be found at: https://youtu.be/saX9zc9pbps}, the Randomized State \gls{RL} agent exhibited smoother behavior compared to the Ideal State agent. It maintained minimal velocities when approaching the object, while the Ideal State agent displayed a jittery movement pattern due to operating at maximum velocity ranges, resulting in overshooting. This highlights the importance of incorporating observation noise during policy training for numerical agents to ensure robust sim2real transfer.

\setlength{\tabcolsep}{1.5mm}
\begin{table}[t]
    \centering
    \caption{Mean success rate across the different state representation strategies.} 
    \label{table:experiment2Results}
    \begin{tabular}{p{11mm} cccc@{}}
    \hline
       &             \specialcell{Average + Std.\\(Idl. Sim.)}                  & \specialcell{Best Model\\(Idl. Sim.)} & \specialcell{Best Model\\(Rnd. Sim.)} & \specialcell{Best Model\\(Real)} \vspace{-9pt}\\
Strategy \ & & &\\
    \hline
    Ruckig &  \textbf{100\%} &  \textbf{100\%} & N/A & \textbf{100\%} \\
    St. & \textbf{100\%} &  \textbf{100\%}  &  N/A   & \textbf{100\%} \\
    St. (rnd.) & \textbf{100\%}  &  \textbf{100\%} & N/A  & \textbf{100\%} \\
    VtS  &  91.6\% $\pm$ 2.2 &  94\%  &  70\% & 52\%  \\
    IGAE   &  96.0\% $\pm$ 2.8  &  \textbf{100\%}  &  78\%  & \textbf{84}\% \\
    AE   &   82.4\% $\pm$ 7.1 &  92\%  &  70\%   & 60\% \\
    EtE   &  54.8\% $\pm$ 11.0  &  78\%  &  44\% & 24\% \\
    \hline  
    \end{tabular}
\end{table}

\section{Discussion}

The results show a significant difference in performance between \gls{RL} agents relying on numerical vs. image-based representations. Since solving the task requires high precision, a representation generated from the robot's encoders is more suitable to learn from, compared to a more abstract representation inferred from images. The performance of image-based representations can be improved if we introduce domain knowledge and provide an incentive about the important aspects of the task, as is done with the IGAE approach. Otherwise, a vanilla autoencoder would assign equal importance to all pixels in the input observations. This is a problem in the specific case, because the robot and the object occupy a small portion of the workspace, making the latent vectors more prone to noise. However, general representations like the ones generated from the vanilla autoencoder become crucial in scenarios where searching for suitable objectives (intervention incentives) to shape the original representation space is not straightforward. For instance, in tasks like pushing objects in cluttered scenes or collaborative tasks, where defining precise objectives can be challenging.

The results provide empirical evidence that decoupling the representation learning from the policy learning leads to better performance. In the case of end-to-end learning, both the encoder and policy network weights are backpropagated during the policy update steps, and the only signal to refine the state of the agent is the reward function. In future works, intervention incentives for control, like action primitives \cite{dalal2021raps}, should be considered to reduce the optimization difficulty of the \gls{RL} policy again.

It also seems that even small errors in predicting the state from the \gls{VtS} agent could lead to detrimental results. A minor error of $2^{\circ}$ could result in an average error of 5 cm in task space for a 1.5 m robot and lose the target. Noisy predictions in some of the dimensions of the 128-dim latent vectors of the autoencoders do not affect at the same level the agent behavior. In VtS models, it would be more practical instead to regress the pose of the end-effector, where the predictions could be more precise. This has been ignored in our analysis due to preserving a fair comparison with the state RL agents, which rely on the values of the robot's joints.

It is plausible that fine-tuning the individual networks, such as carefully designing the architectures of the neural networks in \gls{VtS} and end-to-end approaches and optimizing all the hyperparameters of the \gls{PPO} algorithm from the beginning, could lead to higher success rates for the grasping task. However, we focused on ensuring equal conditions as much as possible throughout the approaches to examine the advantages and weaknesses of each approach more objectively. 

The evaluation of the representation quality of the agents that utilize the pre-trained encoder can be done using a Nearest-Neighbor criterion, as in \cite{Lesort2017UnsupervisedSR, LESORT2018379}. First, we must acquire the nearest neighbors in the learned state space, and then, for each neighbor, find the corresponding state of the robot as defined in the case of the Numerical agents \ref{state-numerical}. KNN-MSE measures how far an observation's value is from its nearest neighbors in the learned data, using the ground truth states. A small distance means the neighbor relationships in the original data are maintained in the learned representation.

For an observation image I, KNN-MSE is defined via its associated learned state $s = \phi(I)$ as
\begin{equation}
\text{KNN-MSE}(s) = \frac{1}{k} \sum_{s' \in \text{KNN}(s, k)} \|\tilde{s} - \tilde{s}'\|^2 ,
\end{equation}
where $\text{KNN}(s, k)$ returns the $k$ nearest neighbors of $s$, in the learned state space \textbf{S}, $\tilde{s}$ is the ground truth associated to $s$, and $\tilde{s}'$ is the ground truth associated to ${s}'$.

In real experiments, we captured images during the evaluation of the IGAE agent in several episodes and then used them to evaluate this criterion across the different autoencoder-based agents, namely IGAE, AE, and VtS. We opted for $k=3$ to maintain a suitable balance between Bias, Variance, and Interpretability \cite{bishop2006pattern}. The final results are presented in Table \ref{tab:summary}. The results from the KNN-MSE analysis align with the trend observed in the mean success rate. IGAE exhibits the lowest mean KNN-MSE error, followed by AE, and then Vts.

\begin{table}[t]
\centering
\caption{Evaluation of autoencoder-based vision models over KNN-MSE criterion.}
\begin{tabular}{lllll} 
    \hline  
Strategy & Mean & Std. & Max & Min \\
\hline  
IGAE & \textbf{0.0393} & \textbf{0.1679} & 1.8883 & $\mathbf{1.4122 \times 10^{-6}}$ \\
AE   & 0.0459 & 0.1839 & 1.6960 & $\mathbf{1.4122 \times 10^{-6}}$ \\
VtS  & 0.0488 & 0.1946 & \textbf{1.6857} & $1.8881 \times 10^{-6}$ \\
\hline  
\end{tabular}
\label{tab:summary}
\end{table}

\hspace{2cm}
\section{Conclusion and Future Work}
In this study, we provided practical insights into the complex relationship between state representation in reinforcement learning, available system and environment information, and sim2real transfer. Using a grasping task that is easy to solve for a model-based approach with full system knowledge, we examined how well an \gls{RL} agent that uses a representation with an increasing level of abstraction can solve the same task. While \gls{RL} agents with hand-crafted representation had 100\% success, it was challenging to solve the task from images, considering the high precision needed in combination with the low-resolution image input and the simulation-only training. We found that autoencoder-based methods need to be incentivized with domain knowledge to provide a robust representation for solving the task. An \gls{RL} agent relying on representation from our proposed Incentivized Grasping AutoEncoder could completely solve the task in simulation and achieve a success rate of 84\% in sim2real transfer. We claim that the additional loss terms in IGAE make feature detection easier in low-resolution frames since the reconstruction is incentivized to focus on the smaller segmented areas of the gripper and object.

In future works, we would like to further explore the potential of autoencoders and other general or incentivized representation learning approaches in complex tasks, where \gls{RL} agents should possess the flexibility to discern the critical aspects of the environment they operate in and assess whether relying on user-defined and hard-coded state representations can constrain the agent's adaptability. Through tasks that require rich contact, like grasping in cluttered scenes, object pushing, or interaction with novel objects, as well as through architectures like vision transformers, we anticipate gaining deeper insights into the most effective state representation approaches for robotics applications.

\addtolength{\textheight}{-4cm}

\bibliographystyle{IEEEtran}
\bibliography{IEEEabrv,bibliography.bib}
\end{document}